\documentclass[10pt,twocolumn,letterpaper]{article}
\usepackage[table,dvipsnames]{xcolor}

\usepackage{cvpr}              % To produce the CAMERA-READY version
\usepackage{multirow}
\usepackage{bbding}
\usepackage{pifont}
\usepackage{extpfeil}
\usepackage{makecell}
\usepackage{extarrows}

\definecolor{mygray}{RGB}{230, 230, 230}

\definecolor{cvprblue}{rgb}{0.21,0.49,0.74}
\usepackage[pagebackref,breaklinks,colorlinks,allcolors=cvprblue]{hyperref}
\def\paperID{***} % *** Enter the Paper ID here
\def\confName{CVPR}
\def\confYear{2025}

\title{DiscoVLA: Discrepancy Reduction in Vision, Language, and Alignment\\for Parameter-Efficient Video-Text Retrieval}

\author{
\textbf{Leqi Shen}$^{1,2,4}$\footnotemark[1]
\quad
\textbf{Guoqiang Gong}$^{5}$\footnotemark[2]
\quad
\textbf{Tianxiang Hao}$^{1,2}$
\quad
\textbf{Tao He}$^{6,7}$
\quad
\textbf{Yifeng Zhang}$^{5}$
\\
\textbf{Pengzhang Liu}$^{5}$
\quad
\textbf{Sicheng Zhao}$^{2}$\footnotemark[3]
\quad
\textbf{Jungong Han}$^{3}$
\quad
\textbf{Guiguang Ding}$^{1,2}$\footnotemark[3]\\
$^{1}$ School of Software $^{2}$ BNRist $^{3}$ Department of Automation, Tsinghua University \\
$^{4}$ Hangzhou Zhuoxi Institute of Brain and Intelligence  \\
\quad
$^{5}$ JD.com
\quad
$^{6}$ GRG Banking Equipment Co., Ltd.
\quad
$^{7}$ South China University of Technology 
}

\begin{document}
\maketitle

{
\renewcommand{\thefootnote}{\fnsymbol{footnote}}
\footnotetext[1]{lunarshen@gmail.com. Work done during an internship at JD.com. }
\footnotetext[2]{Project leader.\quad$^{\ddagger}$ Corresponding authors.}
}

\begin{abstract}
The parameter-efficient adaptation of the image-text pretraining model CLIP for video-text retrieval is a prominent area of research. While CLIP is focused on image-level vision-language matching, video-text retrieval demands comprehensive understanding at the video level. Three key discrepancies emerge in the transfer from image-level to video-level: vision, language, and alignment. However, existing methods mainly focus on vision while neglecting language and alignment. In this paper, we propose \textbf{Disc}repancy Reducti\textbf{o}n in \textbf{V}ision, \textbf{L}anguage, and \textbf{A}lignment (\textbf{DiscoVLA}), which simultaneously mitigates all three discrepancies. Specifically, we introduce Image-Video Features Fusion to integrate image-level and video-level features, effectively tackling both vision and language discrepancies. Additionally, we generate pseudo image captions to learn fine-grained image-level alignment. To mitigate alignment discrepancies, we propose Image-to-Video Alignment Distillation, which leverages image-level alignment knowledge to enhance video-level alignment. Extensive experiments demonstrate the superiority of our DiscoVLA. In particular, on MSRVTT with CLIP (ViT-B/16), 
DiscoVLA outperforms previous methods by \textbf{1.5}\% in R@1, reaching a final score of \textbf{50.5}\% R@1.
The code is available at \url{https://github.com/LunarShen/DsicoVLA}.
\end{abstract}    
\section{Introduction}
\label{sec:intro}

\begin{figure}[t]
    \centering
    \begin{minipage}[b]{\linewidth}
        \centering
        \scalebox{0.9}
        {
        \includegraphics[width=1.0\linewidth]{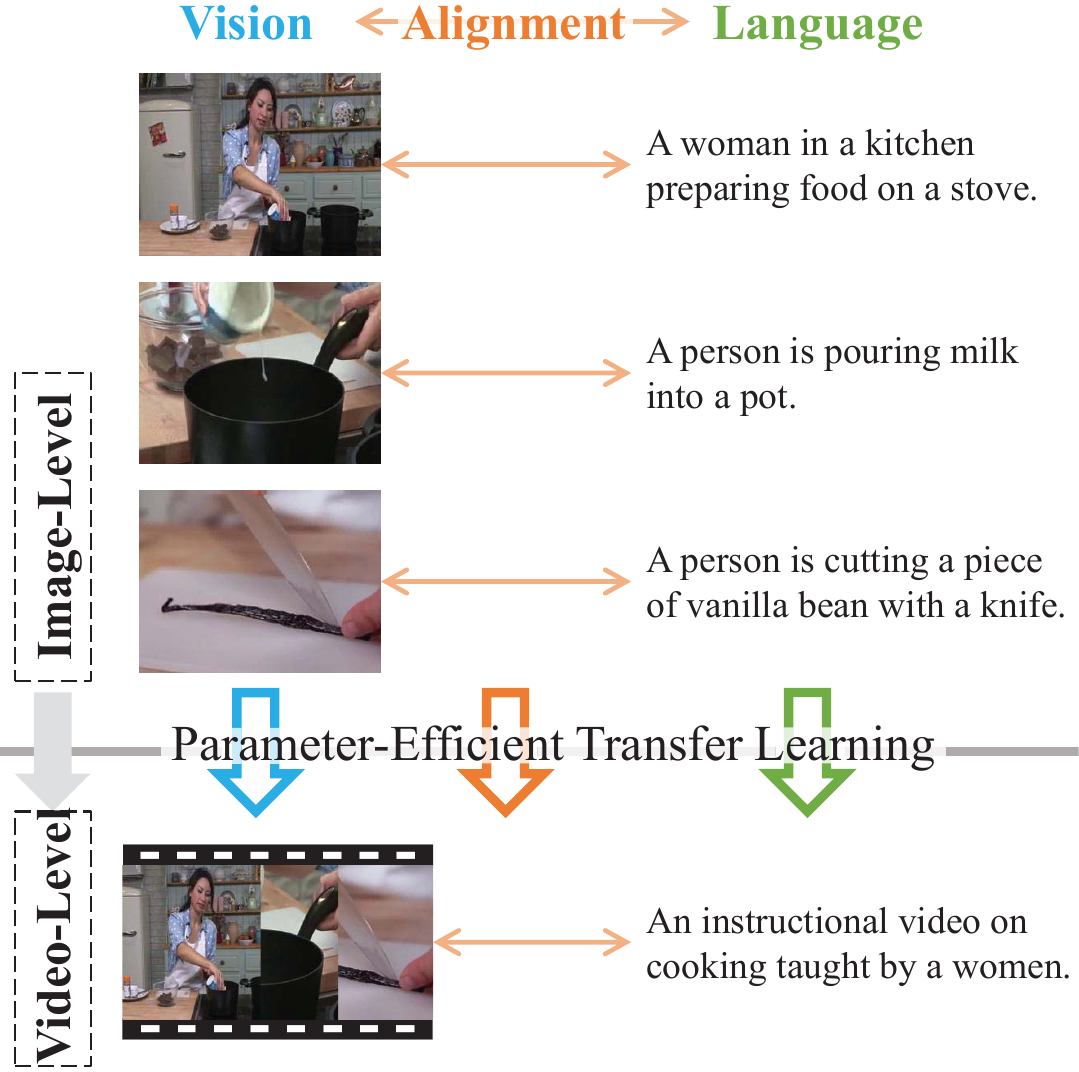}
        }
    \end{minipage}
    
    \vspace{1mm}
    
    \begin{minipage}[b]{\linewidth}
        \centering
        \scalebox{0.8}
        {
        \begin{tabular}{c|ccc}
        \toprule
        % $\stackrel{\rm PETL}{\Longrightarrow}$
        \multirow{2}{*}{Method} & \multicolumn{3}{c}{Image-Level $\xLongrightarrow{\rm PETL}$ Video-Level} \\ \cline{2-4}
                                & Vision   & Language   & Alignment  \\ \hline
        Previous Methods        &  \Checkmark        & \ding{55}           & \ding{55}           \\
        Our DiscoVLA         & \Checkmark         & \Checkmark           & \Checkmark           \\ \bottomrule
        \end{tabular}
        }

    \end{minipage}
    \vspace{-5pt}
    \caption{When parameter-efficient transferring image-level CLIP to video-text retrieval, we identify three key discrepancies: vision, alignment, and language. Unlike previous methods, our DiscoVLA aims to address all these discrepancies. More visualization examples are provided in Figure~\ref{fig:experiment_vis}.}
    \vspace{-5pt}
    \label{fig:intro_3gaps}
\end{figure}

The rapid expansion of online video content has generated a growing demand for effective video-text retrieval, focusing on matching video content with relevant textual descriptions.
Inspired by the success of image-text pretraining approaches such as CLIP~\cite{radford2021learning}, recent efforts~\cite{luo2022clip4clip,gorti2022x,Wang2022DisentangledRL,deng2023prompt,wu2023cap4video,jin2023video,wang2024text} have increasingly focused on extending these powerful capabilities to video-text retrieval tasks. However, fully fine-tuning these pretrained models requires updating numerous parameters for every dataset, resulting in large storage overhead~\cite{huang2023vop, yang2024dgl}. Therefore, we focus on parameter-efficient transfer learning methods, targeting strong performance with minimal trainable parameters.

Although both image-text pretraining and video-text retrieval involve vision-language (VL) matching, they exhibit substantial discrepancies that hinder straightforward knowledge transfer. 
To be specific, image-text pretraining focuses on matching images with image captions, which represents image-level VL matching. 
In contrast, video-text retrieval focuses on matching videos with video captions, which represents video-level VL matching.
The challenges of video-level VL matching, which demands understanding motion and time, create significant obstacles for transferring knowledge from image-text to video-text tasks. 

In this work, we explore the discrepancies between image-level and video-level VL matching across three key aspects: vision, language, and alignment. 
Videos introduce a temporal dimension, which is absent in images.
Similarly, the distinction between image captions and video captions arises from the varying levels of granularity, such as isolated image descriptions versus continuous video narratives. 
Thus, video-level alignment is inherently more complex than image-level alignment, as it must account for intricate spatio-temporal relationships. 
In Figure~\ref{fig:intro_3gaps}, the three images correspond to ``preparing food'', ``pouring'', and ``cutting'', while the entire video demonstrates an instructional cooking process.
Image-level VL matching focuses on isolated images, while video-level VL matching emphasizes the overall understanding of continuous segments.

However, current methods~\cite{huang2023vop,cao-etal-2024-rap,yang2024dgl,jin2024mv} struggle to fully address the discrepancies mentioned above.  
Some~\cite{huang2023vop,cao-etal-2024-rap,yang2024dgl,jin2024mv} focus on the vision aspect, modeling temporal relationships to extract video-level features, while others~\cite{yang2024dgl,jin2024mv} attempt to reduce the modality gap between vision and language through shared parameter mechanisms. 
Despite these advances, these methods remain limited in both language and alignment discrepancies. 

To tackle these challenges, we introduce \textbf{Disc}repancy Reducti\textbf{o}n in \textbf{V}ision, \textbf{L}anguage, and \textbf{A}lignment, termed \textbf{DiscoVLA}, a novel parameter-efficient method for video-text retrieval. 
Unlike previous methods that target a single gap, our method aims to simultaneously reduce all significant discrepancies, as illustrated in the table of Figure~\ref{fig:intro_3gaps}.

Specifically, we propose an Image-Video Features Fusion module (IVFusion), which addresses both vision and language discrepancies through a unified feature extraction approach. IVFusion effectively combines both image- and video-level features by utilizing a lightweight adapter. 
In addition to video-level alignment, we propose Pseudo Image-level Alignment (PImgAlign) which generates pseudo image captions and learns fine-grained image-level alignment.
Furthermore, to reduce alignment discrepancy, we propose Image-to-Video Alignment Distillation (AlignDistill), transferring image-level alignment knowledge to improve video-level alignment.
All proposed modules prioritize parameter efficiency: IVFusion requires minimal trainable parameters, while PImgAlign and AlignDistill operate without increasing inference parameters.

In summary, the main contributions are as follows:
\begin{itemize}
\item We reveal, for the first time, the necessity of addressing all three image-to-video discrepancies---vision, language, and alignment---to achieve parameter-efficient VTR.
\item We introduce IVFusion to fuse image- and video-level features for both vision and language gaps. PImgAlign is introduced for fine-grained image-level alignment. We introduce AlignDistill to minimize alignment gaps.
\item Our DiscoVLA achieves state-of-the-art results on MSRVTT, LSMDC, ActivityNet, and DiDeMo. Notably, on MSRVTT with CLIP (ViT-B/16), DiscoVLA reaches \textbf{50.5}\% R@1, surpassing previous methods by \textbf{1.5}\%.
\end{itemize}

\section{Related Work}
\label{sec:related_work}

\noindent \textbf{Video-Text Retrieval.}
Advances in vision-language pretraining~\cite{wang2022internvideo, xue2022clip, chen2023vast, li2023unmasked, fan2024improving,radford2021learning,xu2023demystifying,li2021align,li2022blip} have achieved significant success across various downstream tasks~\cite{antol2015vqa, xu2015show, karpathy2015deep, he2022secret, shen2024x, ding2021repvgg, wang2024yolov10, yang2024llmi3d, yang2024heie, ding2019acnet}. 
Leveraging powerful CLIP~\cite{radford2021learning}, recent video-text retrieval (VTR) methods~\cite{luo2022clip4clip,gorti2022x,Wang2022DisentangledRL,deng2023prompt,wu2023cap4video,jin2023video,wang2024text} have achieved impressive performance enhancements.
CLIP4Clip~\cite{luo2022clip4clip} is a pioneering CLIP-based method that aggregates image features via mean-pooling or transformers to obtain video features, inspiring subsequent studies.
Additionally, Cap4Video~\cite{wu2023cap4video} utilizes ZeroCap~\cite{tewel2022zerocap} and GPT-2~\cite{radford2019language} to generate video-level captions as supplementary video content during both training and inference. In contrast, our DiscoVLA addresses the image-level and video-level gap by generating image-level captions used only in training to improve video-level alignment.
These full fine-tuning methods impose substantial overhead from large trainable parameters.

\noindent \textbf{Parameter-Efficient Transfer Learning.}
Parameter-efficient methods aim to adapt pre-trained models to new tasks with minimal fine-tuning.
The mainstream methods in natural language processing and image tasks include techniques such as Prompt~\cite{khattak2023maple,jia2022visual,zhou2022learning,zhou2022conditional,khattak2023self}, Adapter~\cite{houlsby2019parameter,he2021towards,chen2022adaptformer,zhang2022neural}, and LoRA~\cite{hu2021lora,zhang2022neural}.
Prompt~\cite{khattak2023maple} modifies the input space by adding learnable tokens. 
Adapter~\cite{houlsby2019parameter} introduces a lightweight bottleneck neural network consisting of up- and down-projection layers with a non-linear activation function.
LoRA~\cite{hu2021lora} aims to update up- and down-projection layers by learning low-rank decomposition matrices.

\begin{figure*}[t!]
  \centering
   \includegraphics[width=\linewidth]{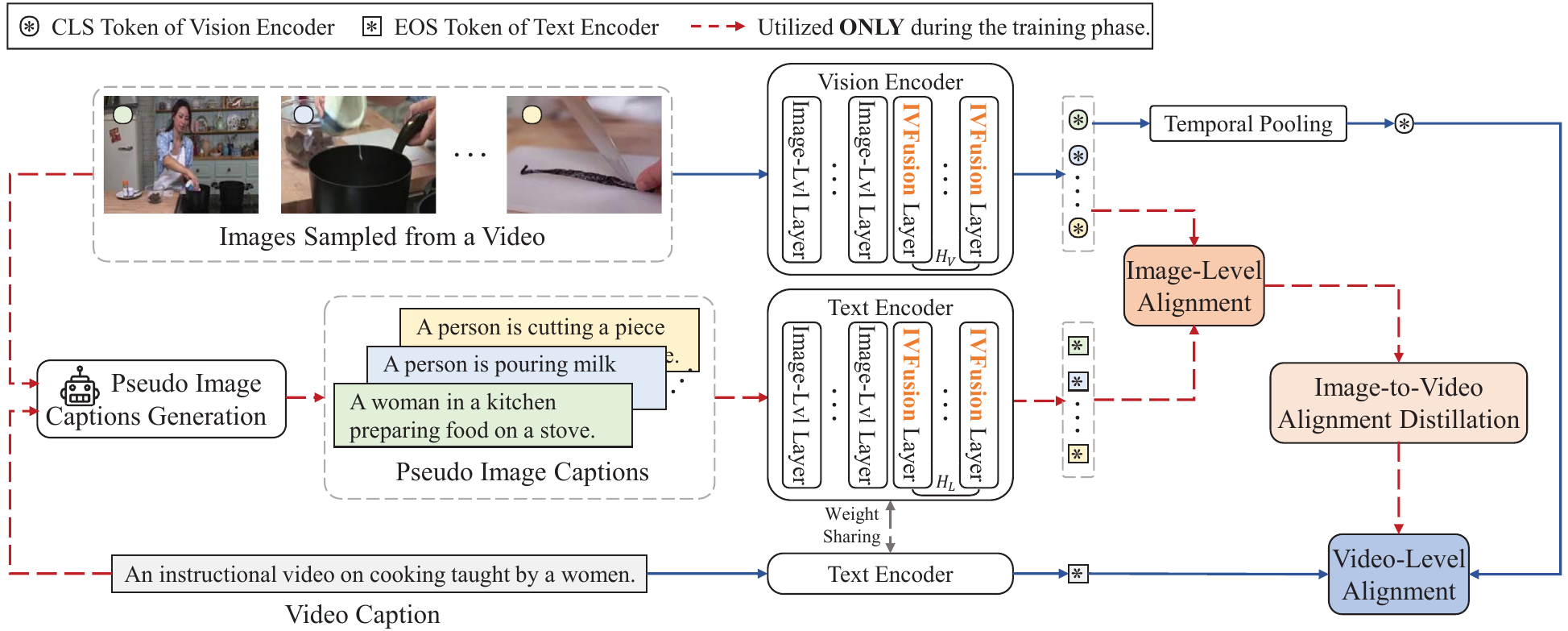}
    \vspace{-22pt}
   \caption{The overall framework of DiscoVLA. Initially, we generate pseudo image captions for each sampled image. In both vision and text encoders, we utilize image-level layers to acquire pretrained image-level knowledge and employ IVFusion layers to enhance spatio-temporal information. 
   The single video caption is encoded through the text encoder, utilizing IVFusion layer as image-level (image-lvl) layer.
   Finally, AlignDistill is applied to distill image-level alignment to video-level alignment. For fair comparisons with previous methods, we do not generate pseudo image captions during the inference phase.}
   \label{fig:method_framework}
\end{figure*}

Recently, several studies have explored parameter-efficient VTR methods~\cite{huang2023vop,cao-etal-2024-rap,yang2024dgl,jin2024mv} to incorporate temporal modeling into the frozen image-level CLIP backbone.
VoP~\cite{huang2023vop} employs trainable BiLSTMs~\cite{graves2005framewise} to produce prompt tokens in the vision encoder.
DGL~\cite{yang2024dgl} introduces weight-sharing layers to generate prompts for both text and video encoders.
MV-Adapter~\cite{jin2024mv} introduces shared bottleneck structures in both video and text branches.
However, these methods overlook language and alignment discrepancies. 
We address these limitations with DiscoVLA, a framework that facilitates robust video-level understanding.
\section{Methodology}
\label{sec:methods}

\subsection{Preliminary}
\label{sec:preliminary}

Video-text retrieval is a task that involves retrieving relevant videos based on a given text query, or conversely, retrieving appropriate text descriptions for a video. Consistent with previous approaches~\cite{huang2023vop,cao-etal-2024-rap,yang2024dgl,jin2024mv}, we utilize the CLIP model, pre-trained on image-text data, as our backbone. 

To process a given video, we first sample a sequence of frame images uniformly, which are then input into the CLIP vision encoder to extract image features $v^{\rm img}$. These features are then aggregated through temporal average pooling to get video features $v^{\rm vid}$.
For a given video caption, it is passed through the CLIP text encoder to extract video caption features $t^{\rm vid}$. Then, we compute the cross-modal contrastive loss~\cite{oord2018representation}, which functions as a video-level alignment term, maximizing the video-level similarity ${\rm Sim}^{\rm vid}$ between matched pairs:
\begin{gather}
{\rm Sim}^{\rm vid}_{ij}=t^{{\rm vid}^T}_{i}v^{\rm vid}_{j}, \label{eq:video_sim} \\
\mathcal{L}_{t2v} ({\rm Sim}^{\rm vid}) = \frac{1}{B} \sum_{i=1}^{B} {\rm log} \frac{{\rm exp}({\rm Sim}^{\rm vid}_{ii})/\tau)}{\sum_{j=1}^B {\rm exp}({\rm Sim}^{\rm vid}_{ij})/\tau)}, \\
\mathcal{L}_{v2t} ({\rm Sim}^{\rm vid}) = \frac{1}{B} \sum_{i=1}^{B} {\rm log} \frac{{\rm exp}({\rm Sim}^{\rm vid}_{ii})/\tau)}{\sum_{j=1}^B {\rm exp}({\rm Sim}^{\rm vid}_{ji})/\tau)}, \\
    \mathcal{L}_{A} ({\rm Sim}^{\rm vid}) = - \frac{1}{2}[\mathcal{L}_{t2v} ({\rm Sim}^{\rm vid}) + \mathcal{L}_{v2t} ({\rm Sim}^{\rm vid})], \label{loss:align}
\end{gather}
\noindent where $B$ is the batch size, $\tau$ is the temperature parameter. 

\begin{figure*}[t]
  \centering
  \scalebox{0.96}
  {
  \includegraphics[width=\linewidth]{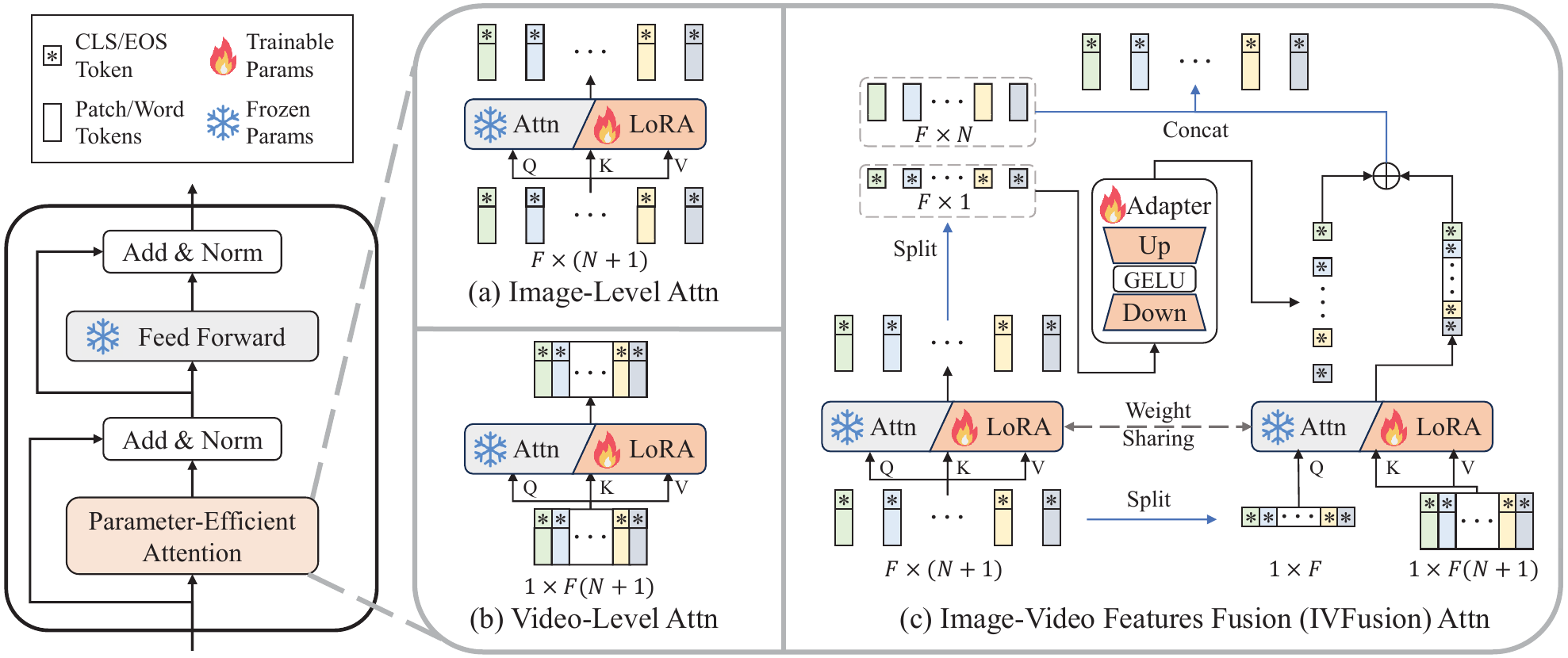}
  }
  \vspace{-10pt}
   \caption{Illustration of the encoder layers for vision and text encoders. (a) Image-Level Attn operates on each of the $F$ images or image captions individually. (b) Video-Level Attn concatenates tokens across all sampled images or image captions. 
   (c) IVFusion Attn employs a lightweight adapter to integrate the efficiency of Image-Level Attn with the effectiveness of Video-Level Attn.
   Here, we illustrate the application of IVFusion within the vision encoder. The text encoder adopts the same approach for processing pseudo image captions.}
   \label{fig:method_IVFusion}
\end{figure*}
 
\subsection{DiscoVLA}

As shown in Figure~\ref{fig:method_framework}, we propose DiscoVLA to address the challenges of vision, language, and alignment when transferring from image-level to video-level VL matching. 
First, we introduce Image-Video Features Fusion (IVFusion) in both vision and text encoders to enhance feature extraction.
Next, we introduce Pseudo Image-Level Alignment (PImgAlign) for fine-grained image-level alignment, consisting of Pseudo Image Language Generation and Image-Level Alignment.
Finally, Image-to-Video Alignment Distillation (AlignDistill) is proposed to leverage image-level alignment knowledge to enhance video-level alignment.

Importantly, PImgAlign is only employed during training. For a fair comparison with previous methods and due to its generation inefficiency, PImgAlign is omitted at inference. The final similarity used for retrieval is based on video-level similarity in Eq.~(\ref{eq:video_sim}).

We focus on parameter-efficient transfer learning for video-text retrieval, where the large backbone parameters remain fixed, and only a small number of additional parameters are trained. The trainable parameters are embedded in the parameter-efficient attention of each encoder layer, as illustrated in Figure~\ref{fig:method_IVFusion}. 

\subsection{Image-Video Features Fusion}

Image-level and video-level variations exist in both vision and language modalities. Our proposed IVFusion fuses these image-level and video-level features in both the vision and language encoders. 
Below, we illustrate IVFusion's application within the vision encoder.

\noindent \textbf{Image-Level Attn.}
To achieve parameter-efficient transfer learning, we apply LoRA in the attention module of each encoder layer. As illustrated in Figure~\ref{fig:method_IVFusion}a, $F\times(N+1)$ tokens are processed via Image-level Attn, where $F$ denotes the number of sampled images, and $N+1$ represents the combination of the global CLS token and patch tokens. Image-Level Attn processes each of the $F$ images individually, using $N+1$ tokens per image. However, Image-Level Attn does not encode temporal information. 

\noindent \textbf{Video-Level Attn.}
A straightforward solution is Video-Level Attn, which concatenates tokens across all sampled images, enabling the attention module to learn spatio-temporal information. As shown in Figure~\ref{fig:method_IVFusion}b, $F(N+1)$ tokens are simultaneously processed during the attention module. Due to the quadratic complexity of the attention mechanism, Image-Level Attn has a complexity of $O(F{(N)}^2)$, while Video-Level Attn has a higher complexity of $O({(FN)}^2)$. 

\noindent \textbf{IVFusion Attn.}
To combine Image-Level Attn’s efficiency with Video-Level Attn’s effectiveness, we introduce IVFusion Attn in Figure~\ref{fig:method_IVFusion}c. We extend Image-Level Attn by incorporating a branch that uses all CLS tokens to extract spatio-temporal information, reducing complexity to $O(F{(N)}^2 + {(F)}^{2}N)$:
\begin{gather}
[c^{(1)}_{i}, {\hat{E}}_{i}]={\rm ATTN}(QKV=[c^{(0)}_{i}, E^{(0)}_{i}]), \\
[c^{(2)}_{1:F}]={\rm ATTN}(Q=[c^{(0)}_{1:F}],KV=[c^{(0)}_{1:F}, E^{(0)}_{1:F}]), \\
{\hat{c}}_{i} = c^{(2)}_{i} + \sigma (c^{(1)}_{i}\mathbf{W}_{\rm down})\mathbf{W}_{\rm up}, \label{eq:ivfusion}
\end{gather}
where $c_i \in \mathbb{R}^{1\times D}$ is CLS token of the $i^{\text{th}}$ image and $c_{1:F} \in \mathbb{R}^{F\times D}$ are CLS tokens of all sampled images. Similarly, $E_i \in \mathbb{R}^{N\times D}$ are patch tokens of the $i^{\text{th}}$ image and $E_{1:F} \in \mathbb{R}^{FN\times D}$ are patch tokens of all sampled images.
In Eq.~(\ref{eq:ivfusion}), we employ a trainable adapter to merge image-level features $c^{(1)}_{i}$ and video-level features $c^{(2)}_{i}$, where $\mathbf{W}_{down} \in \mathbb{R}^{D\times r}$ and $\mathbf{W}_{up} \in \mathbb{R}^{r\times D}$ represent the up- and down-projection layers, and $\sigma$ denotes the non-linear GELU activation~\cite{hendrycks2016gaussian}. Image-level features preserve pretrained spatial details, while video-level features capture temporal information. This fusion mechanism strengthens feature extraction.

Our parameter-efficient vision encoder comprises both image-level and IVFusion layers. The shallow image-level layers employ Image-Level Attn to capture spatial information, while the top $H_v$ IVFusion layers leverage IVFusion Attn to focus on spatio-temporal information. Consequently, our vision encoder outputs enhanced image features $\left\{ v^{\rm img}_{i} | 1 \leq i \leq F\right\} $.

\subsection{Pseudo Image-Level Alignment}

Unlike previous work focused on video-level alignment, we introduce PImgAlign to learn fine-grained image-level alignment via the decomposition of video content at the image level.
To accomplish this, we first generate pseudo image captions in VTR datasets. 
In contrast to the straightforward sampling of images from a video, sampling image captions is impractical, as video captions generally lack the specific details needed for individual images.
As a solution, we utilize the multimodal large language model (MLLM) LLaVA-NeXT~\cite{liu2024llavanext} to generate pseudo image captions.
For each sampled image from a video, we prompt the model using the relevant video caption for guidance:
\begin{quotation}
\noindent The provided image is a frame sampled from the video, which describes $\left\{\textit{video caption}\right\}$. Based on the video’s content, provide a caption for the provided image.
\end{quotation}
\noindent 
Unlike EA-VTR~\cite{ma2024ea}, which uses an image captioner, our method employs a strong MLLM.
By incorporating video caption guidance, it generates higher-quality image captions that align with the overall video content. 
Similarly to visual feature extraction, we adopt our proposed IVFusion to extract pseudo image caption features $\left\{ t^{\rm img}_{i} | 1 \leq i \leq F\right\} $.

Given image features and pseudo image caption features from a paired video and video caption, we introduce a fine-grained image-level similarity. 
Since image captions are typically more concise than the images they describe, a single caption may correspond to multiple images, as shown in Figure~\ref{fig:experiment_vis}d.
Therefore, we calculate the maximum similarity for each image across all image captions and vice versa. 
Specifically, we computed a similarity matrix between image and image caption features, selecting the maximum value from each row and column to obtain the final image-level similarity:
\begin{gather}
    {\rm Sim}^{\rm img}_{ij-t2v} = \frac{1}{F} \sum_{n=1}^{F} \max_{ 1\leq m \leq F} {t^{\rm img}_{i|n}}^{T}{v^{\rm img}_{j|m}}, \\
    {\rm Sim}^{\rm img}_{ij-v2t} = \frac{1}{F} \sum_{m=1}^{F} \max_{ 1\leq n \leq F} {t^{\rm img}_{i|n}}^{T}{v^{\rm img}_{j|m}}, \\
    {\rm Sim}^{\rm img}_{ij}=\frac{1}{2} [{\rm Sim}^{\rm img}_{ij-t2v} + {\rm Sim}^{\rm img}_{ij-v2t}] \label{eq:image_sim},
\end{gather}
where ${\rm Sim}^{\rm img}_{ij}$ denotes the image-level similarity between the $i^{\text{th}}$ video caption and the $j^{\text{th}}$ video.
$t^{\rm img}_{i|n}$ denotes the $n^{\text{th}}$ pseudo image caption features of the $i^{\text{th}}$ video caption 
and 
$v^{\rm img}_{j|m}$ denotes the $m^{\text{th}}$ image features of the $j^{\text{th}}$ video. 
PImgAlign achieves video-level alignment from the image level, effectively emphasizing spatial details.

However, in the inference stage, the lack of ground truth for video captions prevents pseudo image captions generation. Additionally, MLLM introduces a computational overhead. For fair comparisons with other methods, we exclude PImgAlign during inference.

\subsection{Image-to-Video Alignment Distillation}

Pretrained CLIP demonstrates a strong capability for image-level alignment.
In contrast, video-text retrieval relies on comprehensive spatio-temporal alignment. Therefore, we propose AlignDistill to distill image-level alignment knowledge into video-level alignment, thus mitigating alignment discrepancies. Then, we optimize the Kullback-Leibler divergence~\cite{kullback1997information} between image-level similarity and video-level similarity: 
\begin{gather}
\mathcal{S}_{t2v}^{\rm img} = [{\rm s}_{i1}^{\rm img},\cdots,{\rm s}_{iN_{v}}^{\rm img}], \mathcal{S}_{v2t}^{\rm img} = [{\rm s}_{1i}^{\rm img},\cdots,{\rm s}_{N_{t}i}^{\rm img}], \\
\mathcal{S}_{t2v}^{\rm vid} = [{\rm s}_{i1}^{\rm vid},\cdots,{\rm s}_{iN_{v}}^{\rm vid}], \mathcal{S}_{v2t}^{\rm vid} = [{\rm s}_{1i}^{\rm vid},\cdots,{\rm s}_{N_{t}i}^{\rm vid}], \\
\mathcal{L}_{KL} = \frac{1}{2}( \mathbf{KL}(\mathcal{S}_{t2v}^{\rm img} \parallel \mathcal{S}_{t2v}^{\rm vid}) + \mathbf{KL}(\mathcal{S}_{v2t}^{\rm img} \parallel \mathcal{S}_{v2t}^{\rm vid}) ). \label{loss:distill}
\end{gather}
where 
${\rm s}_{ij}^{\rm img}$ is ${\rm Sim}_{ij}^{\rm img}$ in Eq.~(\ref{eq:image_sim})
and
${\rm s}_{ij}^{\rm vid}$ is ${\rm Sim}_{ij}^{\rm vid}$ in Eq.~(\ref{eq:video_sim}).
$\mathcal{S}_{t2v}^{\rm img}$ and $\mathcal{S}_{t2v}^{\rm vid}$ denote the probability distributions for the text-to-video task, while $\mathcal{S}_{v2t}^{\rm img}$ and $\mathcal{S}_{v2t}^{\rm vid}$ denote those for the video-to-text task. 
During the inference phase, We employ enhanced video-level similarity for retrieval tasks.

Finally, the overall training objective, which includes alignment loss (Eq.~(\ref{loss:align})) and distillation loss (Eq.~(\ref{loss:distill})), is formulated as follows:
\begin{gather}
\mathcal{L} = \mathcal{L}_{A} ({\rm Sim}^{\rm vid}) + \alpha \mathcal{L}_{A} ({\rm Sim}^{\rm img}) + \beta  \mathcal{L}_{KL}, \label{eq:all_loss}
\end{gather}
where $\alpha$ and $\beta$ are used to balance the loss. $\mathcal{L}_{A} ({\rm Sim}^{\rm img})$ is the image-level alignment loss (see Eq.~(\ref{loss:align}) and (\ref{eq:image_sim})).

\section{Experiments}
\label{sec:experiments}

\subsection{Experimental Settings}

\begin{table*}[t]
\centering
% \resizebox{\linewidth}{!}
\scalebox{0.83}
{
\begin{tabular}{cccccccccccc}
\toprule
\multirow{2}{*}{Method} & \multicolumn{1}{|c|}{\multirow{2}{*}{\makecell{\# Params\\(M)}}} & \multicolumn{5}{c|}{Text-to-Video}                     & \multicolumn{5}{c}{Video-to-Text} \\ \cline{3-12} 
                         & \multicolumn{1}{|c|}{}                            & R@1$\uparrow$ & R@5$\uparrow$  & R@10$\uparrow$ & R@sum$\uparrow$ & \multicolumn{1}{c|}{MnR$\downarrow$}  & R@1$\uparrow$ & R@5$\uparrow$  & R@10$\uparrow$ & R@sum$\uparrow$ & MnR$\downarrow$  \\ \hline
\multicolumn{12}{c}{CLIP (ViT-B/32)}                                                                                                                                     \\ \hline
\rowcolor{mygray} Full fine-tuning         & \multicolumn{1}{|c|}{123.54}                      & 43.1 & 70.4 & 80.8 & 194.3 & \multicolumn{1}{c|}{16.2} & 43.1 & 70.5 & 81.2 & 194.8 & 12.4 \\
Prompt~\cite{khattak2023maple}                   & \multicolumn{1}{|c|}{0.08}                        & 40.4 & 66.3 & 77.3 & 184.0 & \multicolumn{1}{c|}{16.7} & 42.2 & 69.7 & 79.2 & 191.1 & 12.4 \\
Adapter~\cite{houlsby2019parameter}                 & \multicolumn{1}{|c|}{0.26}                        & 41.9 & 69.9 & 78.7 & 190.2 & \multicolumn{1}{c|}{14.9} & 43.6 & 69.9 & 80.1 & 193.6 & 11.5 \\
LoRA~\cite{hu2021lora}                     & \multicolumn{1}{|c|}{0.49}                        & 43.7 & 68.9 & 80.4 & 193.0 & \multicolumn{1}{c|}{16.0} & 43.0 & 70.2 & 82.2 & 195.4 & 12.0 \\
RAP~\cite{cao-etal-2024-rap}           & \multicolumn{1}{|c|}{1.06}                        & 44.8     & 71.4     & 81.5     & 197.7      & \multicolumn{1}{c|}{14.4}     & 44.0     & 71.9     & 82.4     & 198.3      & 10.1     \\
VoP$^{\rm F+P}$~\cite{huang2023vop}                 & \multicolumn{1}{|c|}{0.4}                         & 43.5 & 69.3 & 79.3 & 192.1 & \multicolumn{1}{c|}{14.8} & 43.6 & 71.2 & 81.2 & 196.0 & 11.0 \\
VoP F+C~\cite{huang2023vop}                  & \multicolumn{1}{|c|}{14.10}                       & 44.7 & 70.5 & 79.2 & 194.4 & \multicolumn{1}{c|}{16.2} & 42.1 & 70.0 & 80.6 & 192.7 & 13.4 \\
DGL~\cite{yang2024dgl}                      & \multicolumn{1}{|c|}{0.83}                        & 44.6 & 69.9 & 80.3 & 194.8 & \multicolumn{1}{c|}{16.3} & 44.5 & 70.7 & 80.6 & 195.8 & 11.5 \\
TempMe~\cite{shen2024tempme} & \multicolumn{1}{|c|}{0.50} & 46.1 & 71.8 & 80.7 & 198.6 & \multicolumn{1}{c|}{14.8} & 45.6 & 72.4 & 81.2 & 199.2 & 10.2 \\
\textbf{DiscoVLA}           & \multicolumn{1}{|c|}{0.56}                        & \textbf{47.0} & \textbf{73.0} & \textbf{82.8} & \textbf{202.8} & \multicolumn{1}{c|}{\textbf{14.1}} & \textbf{47.7} & \textbf{73.6} & \textbf{83.6} & \textbf{204.9} & \textbf{10.0} \\ \hline
\multicolumn{12}{c}{CLIP (ViT-B/16)}                                                                                                                                     \\ \hline
MV-Adapter~\cite{jin2024mv}               & \multicolumn{1}{|c|}{3.6}                         & 46.0 & 72.0 & 82.1 & 200.1 & \multicolumn{1}{c|}{-}    & 45.6 & 74.0 & 83.8 & 203.4 & -    \\
RAP~\cite{cao-etal-2024-rap}                      & \multicolumn{1}{|c|}{1.06}                        & 46.5 & 73.9 & 82.0 & 202.4 & \multicolumn{1}{c|}{12.1} & 45.3 & \textbf{76.4} & 84.8 & 206.5 & 9.1  \\
VoP$^{\rm F+P}$~\cite{huang2023vop}                  & \multicolumn{1}{|c|}{0.4}                         & 47.1 & 72.4 & 81.8 & 201.3  & \multicolumn{1}{c|}{12.9}  & -    & -    & -    & -     & -    \\
VoP$^{\rm F+C}$~\cite{huang2023vop}                  & \multicolumn{1}{|c|}{14.10}                       & 47.7 & 72.4 & 82.2 & 202.3 & \multicolumn{1}{c|}{12.0} & -    & -    & -    & -     & -    \\
DGL~\cite{yang2024dgl}                      & \multicolumn{1}{|c|}{0.83}                        & 48.3 & 71.8 & 80.6 & 200.7 & \multicolumn{1}{c|}{13.4} & 45.7 & 74.0 & 82.9 & 202.6 & 10.9 \\
TempMe~\cite{shen2024tempme} & \multicolumn{1}{|c|}{0.50} & 49.0 & 74.4 & 83.3 & 206.7 & \multicolumn{1}{c|}{\textbf{11.9}} & 47.6 & 75.3 & \textbf{85.4} & 208.3 & 9.0 \\
\textbf{DiscoVLA}           & \multicolumn{1}{|c|}{0.56}                        & \textbf{50.5}     & \textbf{75.6}     & \textbf{83.8}     & \textbf{209.9}      & \multicolumn{1}{c|}{12.1}     & \textbf{49.2}     & 76.0     & 84.7      & \textbf{209.9}      & \textbf{8.6}     \\ \bottomrule
\end{tabular}
}
\vspace{-10pt}
\caption{Comparisons with state-of-the-art methods on MSRVTT. \# Params denotes the number of trainable parameters. $\uparrow$ denotes higher values represent better performance and $\downarrow$ denotes lower values represent better performance. The best value for each metric is highlighted in bold. 
R@sum is defined as the total of R@1, R@5, and R@10.
The gray row represents the fully fine-tuned CLIP4Clip-meanP~\cite{luo2022clip4clip}.}
\vspace{-6pt}
\label{tab:sota_msrvtt}
\end{table*}  

\begin{table*}[t]
\centering
% \resizebox{\linewidth}{!}
\scalebox{0.82}
{
\begin{tabular}{c|c|ccccc|ccccc}
\toprule
\multirow{2}{*}{Method} & \multicolumn{1}{c|}{\multirow{2}{*}{\makecell{\# Params\\(M)}}} & \multicolumn{5}{c|}{Text-to-Video}                     & \multicolumn{5}{c}{Video-to-Text} \\ \cline{3-12} 
                         & \multicolumn{1}{c|}{}                            & R@1$\uparrow$ & R@5$\uparrow$  & R@10$\uparrow$ & R@sum$\uparrow$ & \multicolumn{1}{c|}{MnR$\downarrow$}  & R@1$\uparrow$ & R@5$\uparrow$  & R@10$\uparrow$ & R@sum$\uparrow$ & MnR$\downarrow$  \\ \hline
\rowcolor{mygray} Full fine-tuning         & 123.54                      & 20.7  & 38.9 & 47.2 & 106.8 & 65.3 & 20.6 & 39.4 & 47.5 & 107.5 & 56.7 \\
VoP$^{\rm F+P}$~\cite{huang2023vop}                  & 0.4                         & 20.7  & 40.7 & 49.7 & 111.1 & 59.1 & 21.5 & 40.6 & 50.7 & 112.8 & 50.8 \\
VoP$^{\rm F+C}$~\cite{huang2023vop}                  & 14.10                       & 21.1  & 40.9 & 49.6 & 111.6 & 60.1 & 22.3 & 40.3 & 50.7 & 113.3 & 51.1 \\
DGL~\cite{yang2024dgl}                      & 0.83                        & 21.4  & 39.4 & 48.4 & 109.2 & 64.3 & -    & -    & -    & -     & -    \\
\textbf{DiscoVLA}               & 0.56                        & \textbf{23.6}  & \textbf{42.0} & \textbf{52.3} & \textbf{117.9} & \textbf{52.0} & \textbf{22.8} & \textbf{41.9} & \textbf{51.2} & \textbf{115.9} & \textbf{46.6} \\ \bottomrule
\end{tabular}
}
\vspace{-10pt}
\caption{Comparisons with state-of-the-art methods on LSMDC using CLIP (ViT-B/32).}
\vspace{-6pt}
\label{tab:sota_lsmdc}
\end{table*}  

\begin{table*}[t!]
\centering
% \resizebox{\linewidth}{!}
\scalebox{0.82}
{
\begin{tabular}{c|c|ccccc|ccccc}
\toprule
\multirow{2}{*}{Method} & \multicolumn{1}{c|}{\multirow{2}{*}{\makecell{\# Params\\(M)}}} & \multicolumn{5}{c|}{Text-to-Video}                     & \multicolumn{5}{c}{Video-to-Text} \\ \cline{3-12} 
                         & \multicolumn{1}{c|}{}                            & R@1$\uparrow$ & R@5$\uparrow$  & R@10$\uparrow$ & R@sum$\uparrow$ & \multicolumn{1}{c|}{MnR$\downarrow$}  & R@1$\uparrow$ & R@5$\uparrow$  & R@10$\uparrow$ & R@sum$\uparrow$ & MnR$\downarrow$  \\ \hline
\rowcolor{mygray} Full fine-tuning         & 123.54                      & 40.5  & 72.4 & -    & -     & 7.4  & 42.5 & 74.1 & 85.8 & 202.4 & 6.6  \\
RAP~\cite{cao-etal-2024-rap}                      & 1.06                        & 40.8  & 71.0 & 82.2 & 194.0   & 8.3  & -    & -    & -    & -     & -    \\
VoP$^{\rm F+P}$~\cite{huang2023vop}                  & 0.4                         & 36.1  & 65.5 & 78.5 & 180.1 & 10.9 & 36.3 & 65.9 & 79.2 & 181.4 & 10.1 \\
VoP$^{\rm F+C}$~\cite{huang2023vop}                  & 14.10                       & 35.1  & 63.7 & 77.6 & 176.4 & 11.4 & 35.6 & 65.9 & 77.8 & 179.3 & 10.4 \\
DGL~\cite{yang2024dgl}                      & 0.83                        & 38.6  & 69.2 & 81.6 & 189.4 & 9.0  & -    & -    & -    & -     & -    \\
\textbf{DiscoVLA}           & 0.56                        & \textbf{41.2}  & \textbf{72.4} & \textbf{83.6} & \textbf{197.2} & \textbf{7.8}  & \textbf{41.8} & \textbf{72.8} & \textbf{84.7} & \textbf{199.3} & \textbf{6.9}  \\ \bottomrule
\end{tabular}
}
\vspace{-10pt}
\caption{Comparisons with state-of-the-art methods on ActivityNet using CLIP (ViT-B/32).}
\vspace{-10pt}
\label{tab:sota_anet}
\end{table*} 

Following common practice, we perform evaluations on four widely used benchmarks: MSRVTT~\cite{xu2016msr}, LSMDC~\cite{rohrbach2015long}, ActivityNet~\cite{krishna2017dense}, and DiDeMo~\cite{anne2017localizing}.
We evaluate the performance using common retrieval metrics such as Recall at K (R@K and K = 1, 5, 10), the sum of these recalls (R@sum), and Mean Rank (MnR).
In all experiments, the LoRA dimension and the adapter dimension $r$ are set to 8. In Eq.~\ref{eq:all_loss}, $\alpha$ and $\beta$ are set to $0.3$ and $1.0$, respectively. For the number of IVFusion layers, we set $H_V=4$ for the vision encoder and $H_L=2$ for the text encoder. Please see Appendix~\ref{sec:extra_set} for further details.

\subsection{Comparisons with State-of-the-Art Methods}

We compare our proposed DiscoVLA with state-of-the-art methods on popular benchmarks such as MSRVTT, LSMDC, ActivityNet, and DiDeMo. 

Full fine-tuning refers to CLIP4Clip-meanP~\cite{luo2022clip4clip}, which averages image features along the temporal dimension. All parameter-efficient methods follow this temporal average pooling strategy to obtain the final video features. 
Although Prompt~\cite{khattak2023maple}, Adapter~\cite{houlsby2019parameter}, and LoRA~\cite{hu2021lora} are widely adopted for parameter-efficient image and NLP tasks, they lack effective temporal modeling for video-level understanding. 
In contrast, methods like RAP~\cite{cao-etal-2024-rap}, VoP~\cite{huang2023vop}, DGL~\cite{yang2024dgl}, and MV-Adapter~\cite{jin2024mv} focus on temporal modeling in the vision modality. 

As shown in Table~\ref{tab:sota_msrvtt}, on MSRVTT, our DiscoVLA with CLIP (ViT-B/32) as the backbone achieves \textbf{47.0}\% R@1 in the text-to-video task (t2v) and \textbf{47.7}\% R@1 in the video-to-text task (v2t), significantly outperforming previous methods.
Furthermore, when using CLIP (ViT-B/16), DiscoVLA achieves improvements of \textbf{1.5}\% in R@1, reaching a final score of \textbf{50.5}\% R@1.
The comparison results on LSMDC, ActivityNet, DiDeMo are shown in Table~\ref{tab:sota_lsmdc}-\ref{tab:sota_didemo}. Our method outperforms all existing approaches across these datasets. 
Existing approaches overlook the critical image-to-video language and alignment discrepancies. In contrast, our proposed DiscoVLA addresses these discrepancies across all three areas: vision, language, and alignment.

\begin{table*}[t!]
\centering
% \resizebox{\linewidth}{!}
\scalebox{0.85}
{
\begin{tabular}{c|c|ccccc|ccccc}
\toprule
\multirow{2}{*}{Method} & \multicolumn{1}{c|}{\multirow{2}{*}{\makecell{\# Params\\(M)}}} & \multicolumn{5}{c|}{Text-to-Video}                     & \multicolumn{5}{c}{Video-to-Text} \\ \cline{3-12} 
                         & \multicolumn{1}{c|}{}                            & R@1$\uparrow$ & R@5$\uparrow$  & R@10$\uparrow$ & R@sum$\uparrow$ & \multicolumn{1}{c|}{MnR$\downarrow$}  & R@1$\uparrow$ & R@5$\uparrow$  & R@10$\uparrow$ & R@sum$\uparrow$ & MnR$\downarrow$  \\ \hline
\rowcolor{mygray} Full fine-tuning         & 123.54                      & 43.4  & 70.2 & 80.6 & 194.2 & 17.5 & 42.5 & 70.6 & 80.2 & 193.3 & 11.6 \\
RAP~\cite{cao-etal-2024-rap}                      & 1.06                        & 42.6  & 70.4 & 79.6 & 192.6 & 18.0 & -    & -    & -    & -     & -    \\
VoP$^{\rm F+P}$~\cite{huang2023vop}                  & 0.4                         & 45.3  & 72.3 & 80.4 & 198.0 & 13.8 & 44.7 & 71.2 & 81.1 & 197.0 & 9.9  \\
VoP$^{\rm F+C}$~\cite{huang2023vop}                  & 14.10                       & 46.4  & 71.9 & 81.5 & 199.8 & \textbf{13.6} & 44.4 & 71.8 & 81.8 & 198.0 & 9.5  \\
\textbf{DiscoVLA}               & 0.56                        & \textbf{48.4}  & \textbf{74.5} & \textbf{82.7} & \textbf{205.6} & 14.0 & \textbf{47.7} & \textbf{74.4} & \textbf{83.8} & \textbf{205.9} & \textbf{9.3}  \\ \bottomrule
\end{tabular}
}
\vspace{-10pt}
\caption{Comparisons with state-of-the-art methods on DiDeMo using CLIP (ViT-B/32).}
\vspace{-7pt}
\label{tab:sota_didemo}
\end{table*}

\begin{table*}[t!]
\centering
\resizebox{\linewidth}{!}{
\begin{tabular}{c|ccc|c|cccc|cccc}
\toprule
\multirow{2}{*}{Methods} & \multicolumn{3}{c|}{Components}  & \multirow{2}{*}{\makecell{\# Params\\(M)}} & \multicolumn{4}{c|}{Text-to-Video} & \multicolumn{4}{c}{Video-to-Text} \\ \cline{2-4} \cline{6-13} 
                         & IVFusion & PImgAlign & AlignDistill &                             & R@1$\uparrow$ & R@5$\uparrow$  & R@10$\uparrow$ & R@sum$\uparrow$   & R@1$\uparrow$ & R@5$\uparrow$  & R@10$\uparrow$ & R@sum$\uparrow$  \\ \hline
Pretraind CLIP           & \ding{55}         & \ding{55}       & \ding{55}             & 0                           & 30.8   & 53.8   & 63.3   & 147.9   & 26.6   & 50.1   & 62.0   & 138.7  \\ \hline
LoRA                     & \ding{55}         & \ding{55}       & \ding{55}             & 0.49                        & 43.7   & 68.9   & 80.4   & 193.0   & 43.0   & 70.2   & 82.2   & 195.4  \\
B1                         & \Checkmark         & \ding{55}       & \ding{55}             & 0.54                        & 46.5   & \textbf{73.2}   & 82.0   & 201.7   & 46.6   & 72.6   & 82.9   & 202.1  \\
B2                         & \Checkmark         & \Checkmark       & \ding{55}             & 0.56                        & \textbf{47.0}   & 73.1   & 81.3   & 201.4   & 45.7   & 72.9   & 82.4   & 201.0  \\
B3                         & \Checkmark         & \ding{55}       & \Checkmark             & 0.54                        & 46.6   & 71.6   & 81.0   & 199.2   & 45.9   & 71.9   & 82.2   & 200.0  \\
\textbf{DiscoVLA}               & \Checkmark         & \Checkmark       & \Checkmark             & 0.56                        & \textbf{47.0}   & 73.0   & \textbf{82.8}   & \textbf{202.8}   & \textbf{47.7}   & \textbf{73.6}   & \textbf{83.6}   & \textbf{204.9}  \\ \bottomrule
\end{tabular}
}
\vspace{-10pt}
\caption{Ablation study on the contribution of each proposed component on MSRVTT using CLIP (ViT-B/32). Pretrained CLIP denotes the zero-shot performance of CLIP without any additional training. Our proposed methods are built upon LoRA. B1, B2, and B3 denote various combinations of our proposed components.}
\vspace{-12pt}
\label{tab:ablation_components}
\end{table*}

\begin{table}[t!]
\centering
\resizebox{\linewidth}{!}{
\begin{tabular}{c|cccc}
\toprule
\multirow{2}{*}{\makecell{Parameter-Efficient \\ Attention}} & \multicolumn{4}{c}{Text-to-Video}               \\ \cline{2-5} 
                                               & R@1$\uparrow$ & R@5$\uparrow$  & R@10$\uparrow$  & R@sum$\uparrow$ \\ \hline
Image-Level Attn                                    & 43.7 & 68.9 & 80.4 & 193.0 \\
Video-Level Attn                                   & 45.7 & 70.9 & 80.8 & 197.4 \\
IVFusion Attn \textit{w/o} Adapter                      & 46.0 & 71.0 & 80.4 & 197.4 \\
\textbf{IVFusion Attn}                                  & \textbf{46.5} & \textbf{73.2} & \textbf{82.0} & \textbf{201.7} \\ \bottomrule
\end{tabular}
}
\vspace{-10pt}
\caption{Ablation study on IVFusion on MSRVTT using CLIP (ViT-B/32). We compare several implementations against our proposed IVFusion for Parameter-Efficient Attention (see Figure~\ref{fig:method_IVFusion}). Image-Level Attn represents the LoRA baseline.}
\vspace{-7pt}
\label{tab:ablation_ivfusion}
\end{table}

\begin{table}[t!]
\centering
\resizebox{\linewidth}{!}{
\begin{tabular}{c|cccc}
\toprule
\multirow{2}{*}{\makecell{Similarity \\ in PImgAlign}} & \multicolumn{4}{c}{Text-to-Video}               \\ \cline{2-5} 
                                     & R@1$\uparrow$ & R@5$\uparrow$  & R@10$\uparrow$  & R@sum$\uparrow$ \\ \hline
None                    & 46.6 & 71.6 & 81.0 & 199.2 \\
Video-Level                   & 46.3 & 72.3 & 81.2 & 199.8 \\
Paired Image-Level & 46.1 & \textbf{73.2} & 82.4 & 201.7 \\
\textbf{Fine-grained Image-Level}                     & \textbf{47.0} & 73.0 & \textbf{82.8} & \textbf{202.8} \\ \bottomrule
\end{tabular}
}
\vspace{-10pt}
\caption{Ablation study on similarity in PImgAlign on MSRVTT using CLIP (ViT-B/32). None represents DiscoVLA without image-level alignment optimization, shown as B3 in Table~\ref{tab:ablation_components}. Paired Image-Level is based on matched image-caption pairs. Fine-grained Image-Level represents the full DiscoVLA.}
\vspace{-10pt}
\label{tab:ablation_pimgalign}
\end{table}

\subsection{Ablation Study}

\noindent \textbf{Ablation study on individual components.}
We evaluate the proposed components in Table~\ref{tab:ablation_components}. LoRA applies Image-Level Attn (see Figure~\ref{fig:method_IVFusion}a) in both the vision and text encoders. 
In B1, IVFusion significantly enhances performance by \textbf{8.7}\% R@sum in t2v and \textbf{6.7}\% R@sum in v2t, effectively integrating image-level and video-level features.  
In B2 and B3, we find that using PImgAlign or AlignDistill independently yields limited improvements. 
This is largely due to the notable gap between image-level and video-level (see Figure~\ref{fig:intro_3gaps}).
Image-level PImgAlign struggles to impact video-level tasks, while AlignDistll lacks sufficient image-specific learning. 
When combined, however, they complement each other.
PImgAlign captures fine-grained image-level alignments, and AlignDistill effectively transfers this knowledge to the video-level alignment. 
This improves \textbf{1.1}\% R@sum in t2v and \textbf{2.8}\% R@sum in v2t over B1.
Finally, our full DiscoVLA improves \textbf{9.8}\% R@sum in t2v and \textbf{9.5}\% R@sum in v2t over LoRA.

~

\noindent \textbf{Analysis of trainable parameters.} 
The trainable parameters of our DiscoVLA consist of the LoRA and Adapter parts. 
As shown in Table~\ref{tab:ablation_components}, LoRA in both the vision and text encoders contain $\sim$0.49M parameters. In B1, an additional $\sim$0.05M parameters are introduced by the Adapter in the $H_V=4$ IVFusion layers of the vision encoder. In B2, an additional $\sim$0.02M parameters are introduced by the Adapter in the $H_L=2$ IVFusion layers of the text encoder, which are utilized for pseudo-image caption features in PImgAlign. Overall, the total number of trainable parameters of DiscoVLA is $\sim$0.56M.

\begin{figure*}[t!]
  \centering
   \includegraphics[width=\linewidth]{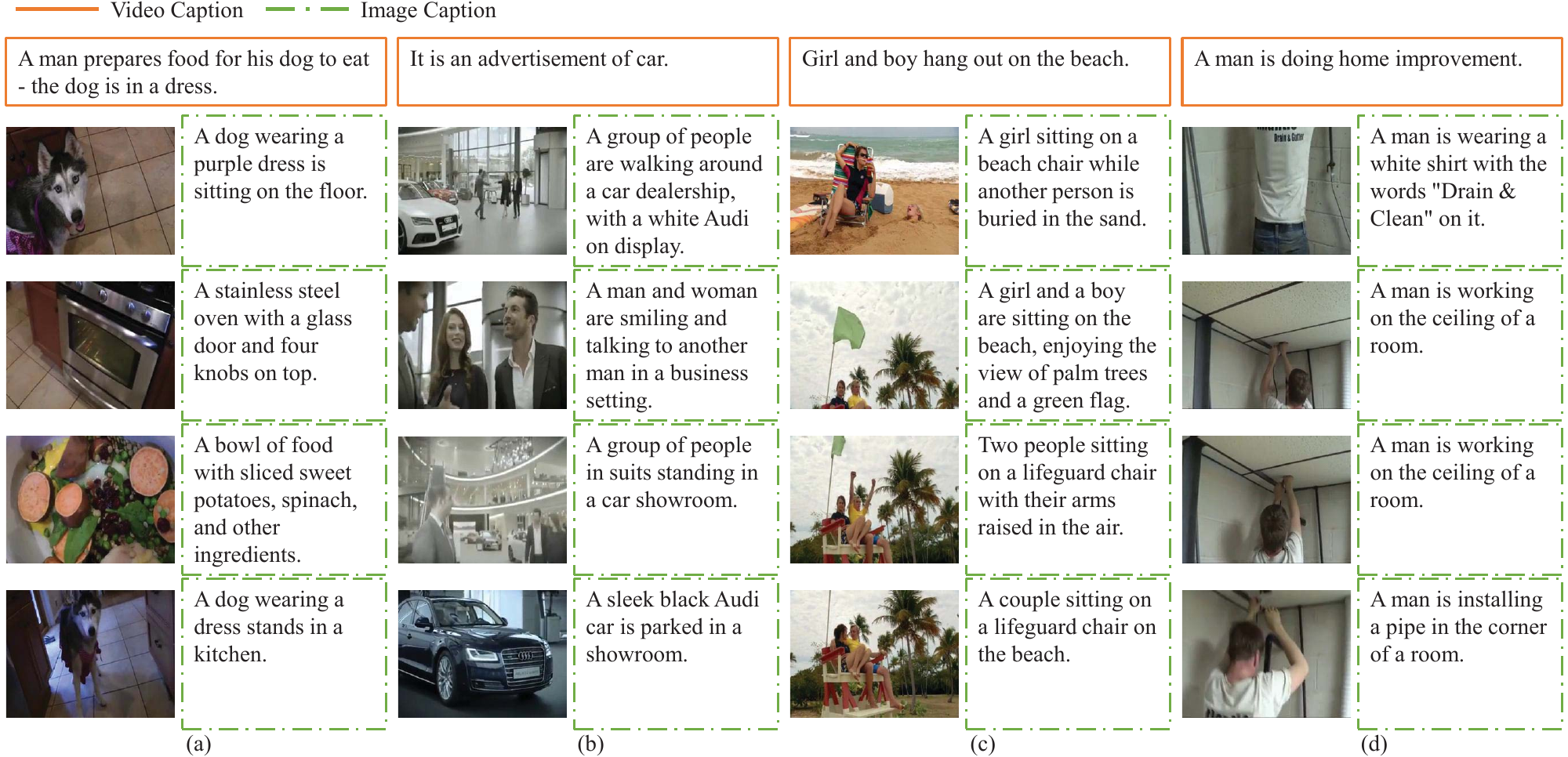}
    \vspace{-20pt}
   \caption{Visualization of discrepancies between video-level and image-level data. Each example consists of a paired video and video caption, with four sampled images from the video. Video captions are highlighted with orange solid lines, and pseudo image captions generated by LLaVA-NeXT~\cite{liu2024llavanext} are indicated with green dashed lines. While video captions convey the general context of the video, image captions focus on the detailed context of each individual frame.}
   \vspace{-3pt}
   \label{fig:experiment_vis}
\end{figure*}

\begin{figure}[t!]
  \centering
   \includegraphics[width=\linewidth]{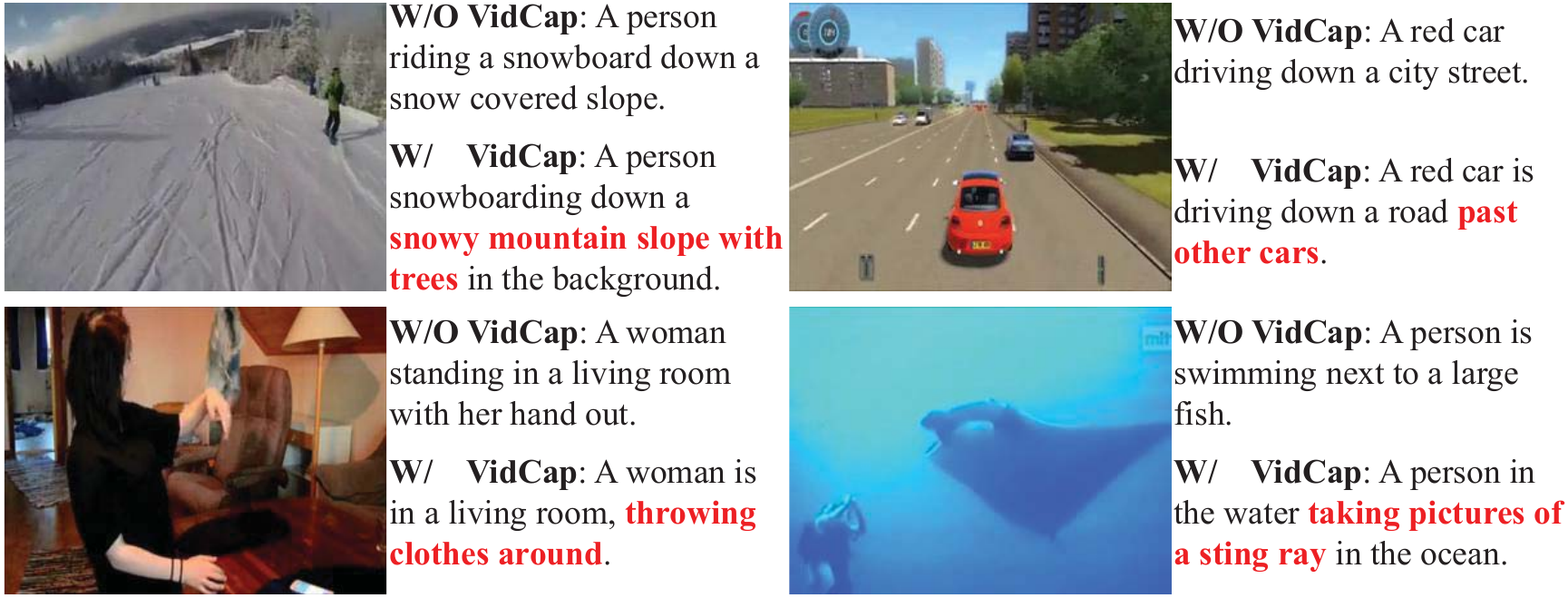}
    \vspace{-20pt}
   \caption{Comparison of image caption generation with (W/) and without (W/O) video caption guidance (VidCap) in PImgAlign.}
   \vspace{-4pt}
   \label{fig:VidCap_Examples}
\end{figure}

\begin{table}[t!]
\centering
\resizebox{\linewidth}{!}{
\begin{tabular}{c|cccc}
\toprule
\multirow{2}{*}{\makecell{Image Caption\\Generation}} & \multicolumn{4}{c}{Text-to-Video}               \\ \cline{2-5} 
                                               & R@1$\uparrow$ & R@5$\uparrow$  & R@10$\uparrow$  & R@sum$\uparrow$ \\ \hline
DiscoVLA   & 47.0                    & 73.0                    & 82.8                     & 202.8 \\
W/O VidCap & 46.1 (-0.9)             & 73.0             & 82.1 (-0.7)              & 201.2 (-1.6) \\ 
\bottomrule
\end{tabular}
}
\vspace{-10pt}
\caption{Ablation study on image caption generation on MSRVTT using CLIP (ViT-B/32). W/O VidCap denotes generating pseudo image captions without video caption guidance.}
\vspace{-10pt}
\label{tab:ablation_vidcap}
\end{table}

\noindent \textbf{Ablation study on IVFusion.}
In Table~\ref{tab:ablation_ivfusion}, we compare different strategies for parameter-efficient attention in Figure~\ref{fig:method_IVFusion}. Video-Level Attn concatenates all tokens across sampled images to capture spatio-temporal information, achieving a notable \textbf{4.4}\% R@sum improvement. To reduce the computational complexity, IVFusion applies spatio-temporal attention only to global CLS tokens and merges image-level features with video-level features. Without an adapter, averaging image- and video-level features results in a modest performance increase. In contrast, with a learnable adapter, IVFusion merges the features more effectively, achieving \textbf{201.7}\% R@sum.

\noindent \textbf{Ablation study on similarity in PImgAlign.}
Table~\ref{tab:ablation_pimgalign} demonstrates the effectiveness of our proposed fine-grained image-level similarity in PImgAlign. 
The similarity is computed between sampled images and pseudo image captions from a paired video and video caption.
The Video-Level baseline averages the image features and pseudo-image caption features along the temporal dimension to compute the video-level similarity, replacing fine-grained image-level similarity. However, this modification leads to a notable decrease in performance, emphasizing the importance of image-level similarity in PImgAlign. 
In contrast, we introduce the Paired Image-Level baseline, which computes similarity based on matched image-caption pairs. Compared to Video-Level, Paired Image-Level yields a \textbf{1.9}\% gain in R@sum.
Moreover, we observe in Figure~\ref{fig:experiment_vis}d that a single image caption may describe multiple images. Consequently, our fine-grained image-level similarity, as defined in Eq.~\ref{eq:image_sim}, selects the maximum similarity from all available pairs, further enhancing video-level alignment.

\noindent \textbf{Ablation study on image caption generation in PImgAlign.}
Figure~\ref{fig:VidCap_Examples} and Table~\ref{tab:ablation_vidcap}
show qualitative and quantitative evaluations of video caption guidance.
In Figure~\ref{fig:VidCap_Examples}, video captions serve as the global context that enriches the understanding of individual frames (marked in red). 
In Table~\ref{tab:ablation_vidcap}, without this guidance, image captions focus only on salient regions while missing globally relevant details, leading to performance drops.
In PImgAlign, we employ MLLM to process both frame images and video descriptions, generating image captions that align with the overall video content.
These high-quality captions further enhance DiscoVLA’s ability to learn robust image-level alignment.

\noindent \textbf{More Ablation Analysis.}
In Appendix~\ref{sec:extra_exp}, we conduct additional ablation studies.
Specifically,
we evaluate the effects of hyperparameters $\alpha,\beta,H_V$ and $H_V$.
For fairness, all results in the main paper are reported without post-processing.
Complementary evaluations with Q-Norm~\cite{bogolin2022cross} and DSL~\cite{cheng2021improving} as post-processing methods are provided in the appendix.

\subsection{Qualitative Analysis}
To compare image-level and video-level data, we conduct a visualization experiment in Figure~\ref{fig:experiment_vis}. 
In Figure~\ref{fig:experiment_vis}a and \ref{fig:experiment_vis}b, image-level data alone fails to capture the complete content of the video. 
Figure~\ref{fig:experiment_vis}c and \ref{fig:experiment_vis}d demonstrate a strong correlation between the video-level and image-level content, but differences across images lead to varying image captions. 
Visually, a video is composed of a sequence of images, incorporating temporal information. Textually, image captions are specific to individual images, while video captions describe the overall video content. 
These discrepancies hinder the effective transfer of image-level CLIP. Our approach addresses this issue by utilizing image-level alignment to enhance video-level alignment.
\section{Conclusion}
In this paper, we identify three major discrepancies in the parameter-efficient transfer of image-level CLIP to video-text retrieval: vision, language, and alignment. To address these challenges, we propose DiscoVLA. Specifically, IVFusion fuses image-level and video-level features to tackle the vision and language discrepancies. For alignment discrepancies, we propose PImgAlign and AlignDistill, which learn and distill image-level alignment to enhance video-level alignment. Our DiscoVLA significantly outperforms existing methods, achieving state-of-the-art performance.
\section*{Acknowledgment}

This work was supported by 
Beijing Natural Science Foundation (No. L223023), 
National Natural Science Foundation of China (Nos. 62441235, 62021002, 62441614), 
the Key R \& D Program of Xinjiang, China (2022B01006), 
Zhejiang Provincial Natural Science Foundation of China under Grant (No. LDT23F01013F01), 
CCF-DiDi GAIA Collaborative Research Funds, 
China Postdoctoral Science Foundation (2024M750565),
Guangdong S \& T Program (2024B0101040008), 
and the Key Realm Research and Development Program of Guangzhou (No. 2024B01W0007).

% \clearpage
{
    \small
    \bibliographystyle{ieeenat_fullname}
    \bibliography{main}
}

\appendix
\clearpage
\maketitlesupplementary

\section{Additional Experimental Settings}
\label{sec:extra_set}

\noindent \textbf{Datasets.}
Following common practice, we perform evaluations on four widely used benchmarks for video-text retrieval: 
(1) \textbf{MSRVTT}~\cite{xu2016msr} includes 10,000 YouTube videos, each with 20 text descriptions. Following previous methods~\cite{gabeur2020multi, miech2019howto100m}, we train on the `train+val' set with 9,000 video-text pairs and evaluate on the `1K-A' test set with 1,000 video-text pairs.
(2) \textbf{LSMDC}~\cite{rohrbach2015long} consists of 118,081 movie clips, each paired with a single description. We use 101,079 for training, 7,408 for validation, and 1,000 for testing, reporting results based on the test set.
(3) \textbf{ActivityNet}~\cite{krishna2017dense} consists of 20,000 YouTube videos. Our evaluation utilizes the `val1' split, which includes 10,009 videos for training and 4,917 for testing. Following previous methods~\cite{gabeur2020multi, zhang2018cross}, we concatenate all sentence descriptions of a video into a single paragraph.
(4) \textbf{DiDeMo}~\cite{anne2017localizing} comprises 10,000 videos with a total of 40,000 text descriptions. The training set contains 8,395 videos, while the test set contains 1,004 videos. Following previous methods~\cite{bain2021frozen,lei2021less}, we combine all descriptions of a video into a single query.

\noindent \textbf{Evaluation Metrics.}
We evaluate the performance using common retrieval metrics such as Recall at K (R@K and K = 1, 5, 10), the sum of these recalls (R@sum), and Mean Rank (MnR). R@K measures the proportion of relevant items retrieved in the top K results for a given query. MnR calculates the mean rank of correct items. Note that for R@K, a higher score means better performance. Conversely, for MnR, a lower score indicates better results.

\noindent \textbf{Implementation Details.}
Following previous parameter-efficient research~\cite{huang2023vop,cao-etal-2024-rap,yang2024dgl,jin2024mv}, we utilize the pre-trained CLIP model as our backbone. we implement the AdamW optimizer~\cite{loshchilov2016sgdr} with a batch size of 128. For all datasets, the initial learning rate is set to 6e-4, employing a cosine learning rate schedule~\cite{goyal2017accurate} over 5 epochs. For MSRVTT and LSMDC, the max frame and caption length are set to 12 and 32. For ActivityNet and DiDeMo, the max frame and caption length are set to 32 and 64. In all experiments, the LoRA dimension and the adapter dimension $r$ are set to 8. In Eq.~\ref{eq:all_loss}, $\alpha$ and $\beta$ are set to $0.3$ and $1.0$, respectively. For the number of IVFusion layers, we set $H_V=4$ for the vision encoder and $H_L=2$ for the text encoder.

\section{Additional Experimental Results}
\label{sec:extra_exp}

\begin{figure}[t]
    \centering
    \scalebox{0.97}
    {
    \includegraphics[width=0.95\linewidth]{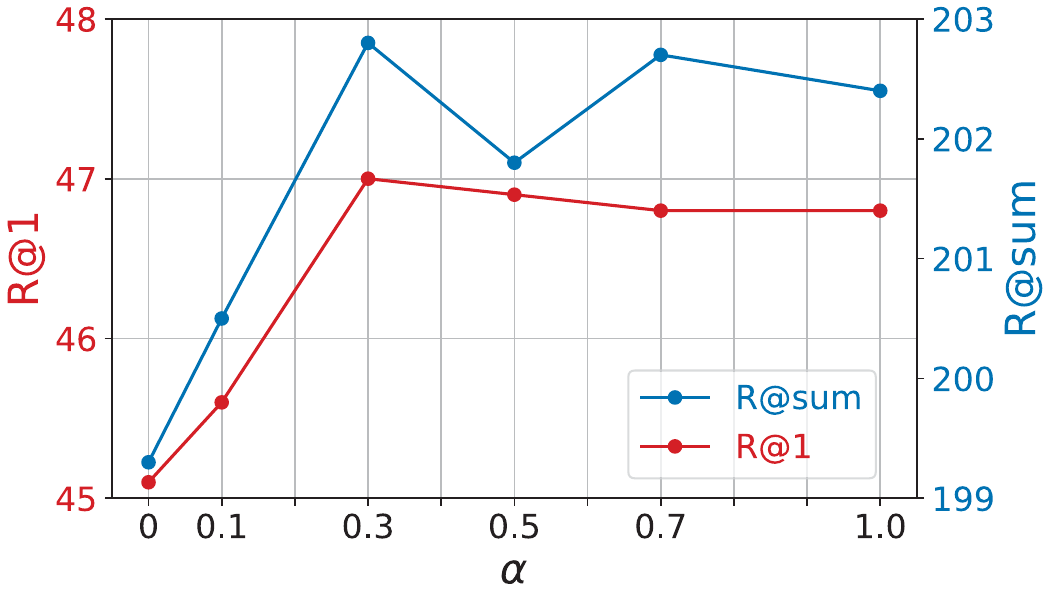} 
    }
    \vspace{-10pt}
    \caption{Ablation study on $\alpha$ in Eq. (\ref{eq:all_loss}) for text-to-video results on MSRVTT using CLIP (ViT-B/32). $\alpha$ represents the weight of the image-level alignment loss $\mathcal{L}_{A}({\rm Sim}^{\rm img})$. All other hyperparameters are kept constant.}
    \label{fig:ablation_imageloss}
\end{figure}

\begin{figure}[t]
    \centering
    \scalebox{0.97}
    {
    \includegraphics[width=0.95\linewidth]{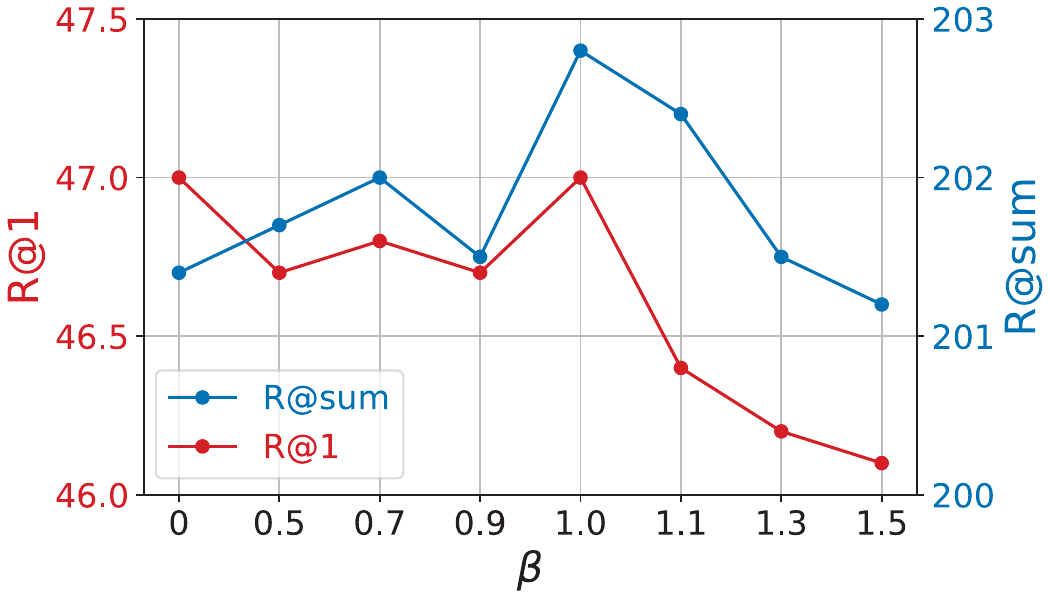} 
    }
    \vspace{-10pt}
    \caption{Ablation study on $\beta$ in Eq. (\ref{eq:all_loss}) for text-to-video results on MSRVTT using CLIP (ViT-B/32). $\beta$ represents the weight of the distillation loss $\mathcal{L}_{KL}$. All other hyperparameters are kept constant.}
    \label{fig:ablation_distllloss}
\end{figure}

\noindent \textbf{Ablation study on $\alpha$ and $\beta$ in Eq. (\ref{eq:all_loss}).}
Figures~\ref{fig:ablation_imageloss} and \ref{fig:ablation_distllloss} present ablation studies on hyperparameters $\alpha$ and $\beta$, respectively. These parameters control the trade-off among the video-level alignment loss $\mathcal{L}_{A} ({\rm Sim}^{\rm vid})$, the image-level alignment loss $\mathcal{L}_{A} ({\rm Sim}^{\rm img})$, and the distillation loss $\mathcal{L}_{KL}$. 
The parameter $\alpha$ represents the weight of $\mathcal{L}_{A}({\rm Sim}^{\rm img})$. Increasing $\alpha$ improves both R@1 and R@sum significantly. Our DiscoVLA achieves optimal performance at $\alpha = 0.3$, beyond which it demonstrates parameter insensitivity. 
The parameter $\beta$ represents the weight of $\mathcal{L}_{KL}$. While $\mathcal{L}_{KL}$ does not affect R@1, setting $\beta = 1.0$ yields a marked improvement in R@sum.
Consequently, we set $\alpha = 0.3$ and $\beta = 1.0$ in our final implementation.

\begin{figure}[t]
    \centering
    \scalebox{0.97}
    {
    \includegraphics[width=0.95\linewidth]{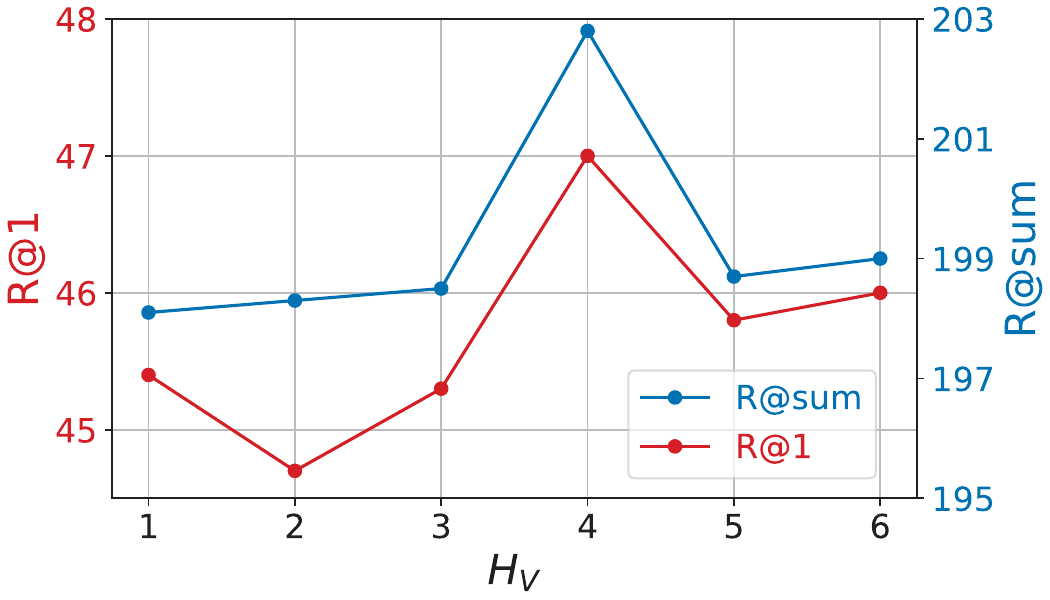} 
    }
    \vspace{-10pt}
    \caption{Ablation study on $H_V$ for text-to-video results on MSRVTT using CLIP (ViT-B/32). $H_V$ represents the number of IVFusion layers in the vision encoder. All other hyperparameters are kept constant.}
    \label{fig:ablation_HV}
\end{figure}

\begin{figure}[t]
    \centering
    \scalebox{0.97}
    {
    \includegraphics[width=0.95\linewidth]{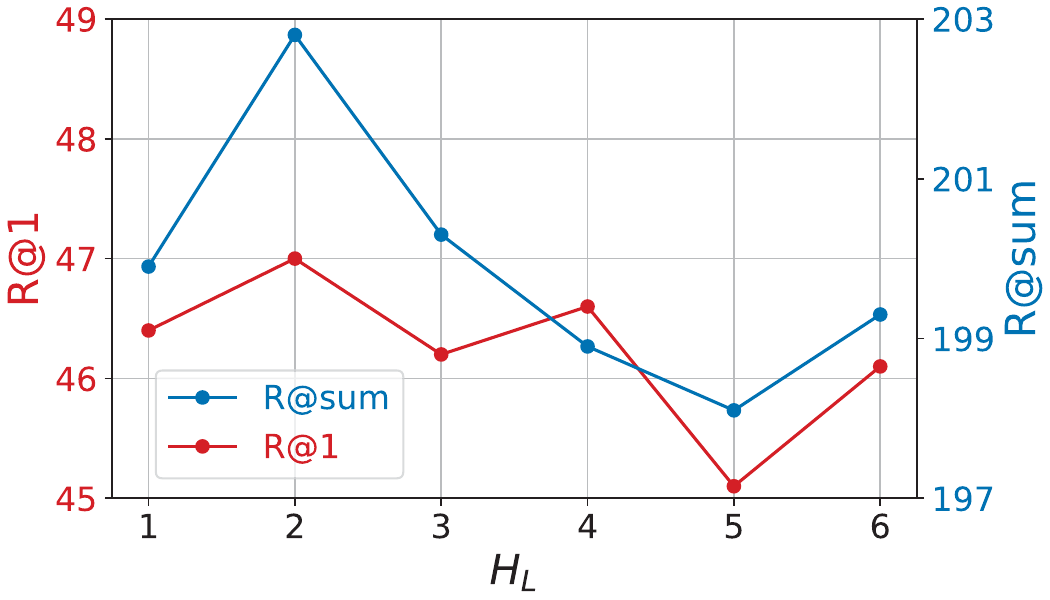} 
    }
    \vspace{-10pt}
    \caption{Ablation study on $H_L$ for text-to-video results on MSRVTT using CLIP (ViT-B/32). $H_L$ represents the number of IVFusion layers in the text encoder. All other hyperparameters are kept constant.}
    \label{fig:ablation_HL}
\end{figure}

\noindent \textbf{Ablation study on the number of IVFusion layers $H_V$ and $H_L$.}
The number of IVFusion layers in the vision and text encoders is denoted by $H_V$ and $H_L$, respectively. 
IVFusion is applied to the upper layers of CLIP, as these layers extract high-level semantic information essential for cross-frame learning.
This learning process requires an adequate number of IVFusion layers.
From Figures \ref{fig:ablation_HV} and \ref{fig:ablation_HL}, we observe that the best performance is achieved when $H_V = 4$ and $H_L = 2$. This difference in optimal values may be attributed to the inherent distinctions between the vision and text modalities. This configuration $H_V = 4$ and $H_L = 2$ demonstrate superior and robust performance across all benchmarks (see Tables~\ref{tab:sota_msrvtt}-\ref{tab:sota_didemo}).

% \usepackage{multirow}
\begin{table}[t]
\centering
\resizebox{\linewidth}{!}
{
\begin{tabular}{cc|cccc}
\toprule
\multicolumn{2}{c|}{\multirow{2}{*}{Post-processing}}                 & \multicolumn{4}{c}{Text-to-Video}                                                                        \\ \cline{3-6} 
 &                                  & \multicolumn{1}{c}{R@1} & \multicolumn{1}{c}{R@5} & \multicolumn{1}{c}{R@10} & \multicolumn{1}{c}{R@sum} \\ \hline
\multicolumn{1}{c|}{\multirow{3}{*}{MSRVTT}} & DiscoVLA   & 47.0                    & 73.0                    & 82.8                     & 202.8                     \\
\multicolumn{1}{c|}{}                     & +QB-Norm~\cite{bogolin2022cross}   & 47.5            & 73.6             & 82.9              & 204.0              \\
\multicolumn{1}{c|}{}                     & +DSL~\cite{cheng2021improving}   & \textbf{51.3}            & \textbf{77.1}             & \textbf{85.5}              & \textbf{213.9}              \\ \hline
\multicolumn{1}{c|}{\multirow{3}{*}{ActivityNet}} & DiscoVLA   & 41.2                    & 72.4                    & 83.6                     & 197.2                     \\
\multicolumn{1}{c|}{}                     & +QB-Norm~\cite{bogolin2022cross}   & 45.1            & 74.9             & 85.2              & 205.2              \\
\multicolumn{1}{c|}{}                     & +DSL~\cite{cheng2021improving}   & \textbf{49.9}            & \textbf{78.8}             & \textbf{88.1}              & \textbf{216.8}              \\
\bottomrule
\end{tabular}
}
\caption{Effect of post-processing on MSRVTT and ActivityNet using CLIP (ViT-B/32).}
\label{tab:appendix_postprocessing}
\end{table}

\noindent \textbf{Effect of post-processing.}
As shown in Table~\ref{tab:appendix_postprocessing}, we evaluate post-processing methods Q-Norm~\cite{bogolin2022cross} and DSL~\cite{cheng2021improving} on top of our proposed DiscoVLA. 
While Q-Norm leads to a notable \textbf{8.0}\% R@sum increase on ActivityNet, it has minimal effect on MSTVTT. In contrast, DSL provides consistent performance gains across both datasets, improving performance by \textbf{11.1}\% R@sum on MSRVTT and \textbf{19.6}\% on ActivityNet.

\end{document}